\documentclass{article}

\PassOptionsToPackage{numbers, compress}{natbib}

\usepackage[preprint]{neurips_2020}

\usepackage[utf8]{inputenc} 
\usepackage[T1]{fontenc}    
\usepackage{hyperref}       
\usepackage{url}            
\usepackage{booktabs}       
\usepackage{amsfonts}       
\usepackage{nicefrac}       
\usepackage{microtype}      

\usepackage{algorithm}
\usepackage{algorithmic}
\usepackage{pgfplots}
\usepackage{pgf-pie}

\usepackage{booktabs}
\usepackage{multirow}
\usepackage{multicol}

\usepackage{bm}
\usepackage{xspace}
\definecolor{dark_green}{rgb}{0.0, 0.5, 0.0}
\newcommand{\ourmethod}{\textsc{Bridge}\xspace}
\newcommand{\sourcecode}{\url{https://github.com/zirui-HIT/Bridge_for_Numerical_Reasoning}}

\usepackage{amsmath}
\newtheorem{proposition}{Proposition}[section]
\newtheorem{proof}{Proof}[section]

\usepackage[framemethod=default]{mdframed}
\newmdenv[
  backgroundcolor=red!05,
  linecolor=quoteborder,
  skipabove=1em,
  skipbelow=0em,
  leftline=true,
  topline=false,
  bottomline=false,
  rightline=false,
  linecolor=red!66,
  linewidth=4pt
]{githubquote}

\title{Exploring Equation as a Better Intermediate Meaning Representation for Numerical Reasoning}

\author{%
  Dingzirui~Wang$^{1}$\footnotemark[1]\thanks{Correspondence to: Dingzirui Wang (\url{dzrwang@ir.hit.edu.cn}).} \quad
  Longxu~Dou$^{1}$ \quad
  Wenbin~Zhang$^{2}$ \quad
  Junyu~Zeng$^{2}$ \quad
  Wanxiang~Che$^{1}$ \quad \\
  $^1$ Harbin Insitute of Technology $^2$ Yunfu Technology (Beijing) Co., Ltd. \\
}

\pgfplotsset{compat=1.18}
\begin{document}
    \maketitle
    
    \begin{abstract}
        Numerical reasoning is vital for natural language processing models to understand and process numerical information in real-world scenarios.
        Most current methods first generate the Intermediate Meaning Representations (IMRs) of questions and then generate answers.
        Current SOTA methods generate programs as IMRs with large language models (LLMs).
        Intuitively, equations have fewer restrictions and closer semantics to the question than programs, leading to higher generation accuracy.
        However, current LLMs generate equations worse than programs, where we assume that the equation data is rare in pre-training data compared to programs.
        So in this paper, we try to use equations as IMRs to solve the numerical reasoning task by addressing two problems:
        \textit{(1)} Theoretically, how to prove that the equation is an IMR with higher generation accuracy than programs;
        \textit{(2)} Empirically, how to improve the generation accuracy of equations with LLMs.
        For the first problem, we propose and prove a proposition to theoretically compare the generation accuracy of different IMRs.
        For the second problem, we present a method called \textbf{B}oosting Numerical \textbf{R}eason\textbf{i}ng by \textbf{D}ecomposing the \textbf{G}eneration of \textbf{E}quations (\ourmethod), which can improve the accuracy of LLMs in generating equations as IMRs by reducing the tendency of generating constant expressions and programs.
        Our method improves the performance by 2.2\%, 0.9\%, and 1.7\% on GSM8K, SVAMP, and Algebra datasets compared to the previous state-of-the-art methods under the single reasoning path setting.
        Our codes and prompts are released in \sourcecode.
    \end{abstract}

    \section{Introduction}
        \begin{figure}[t]
    \centering
    \includegraphics[width=0.8\linewidth]{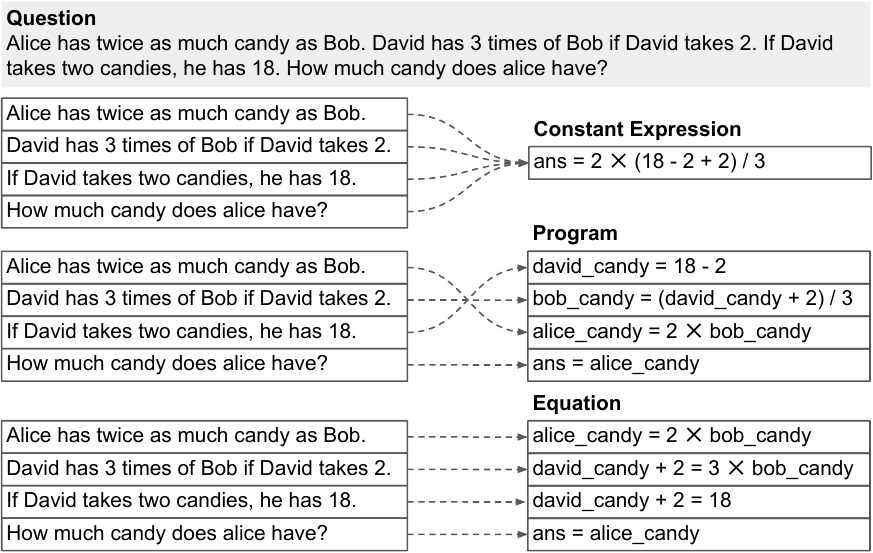}
    \caption{
        Examples of three types of IMRs.
        The dotted line indicates the correspondence between the question and the generated IMR.
        The more complex the correspondence is, the more challenging it becomes to generate accurately.
    }
    \label{fig:intermediate}
\end{figure}

Numerical reasoning is an essential ability of natural language processing (NLP) models to handle documents fulling of numerical information, which is widely used in finance, science, and other fields~\cite{chen-etal-2021-finqa,chen2023theoremqa,lu-etal-2023-survey}.
Generally, numerical reasoning is to generate a value result based on the given question, which describes the values and relationships of quantities~\cite{zhang-etal-2020-graph-tree}.

Numerical calculations, by their inherent complexity, make it a struggle to produce accurate value results directly \cite{thawani-etal-2021-representing}.
To overcome this challenge, most current methods first generate \textbf{I}ntermediate \textbf{M}eaning \textbf{R}epresentations (\textbf{IMRs}) of questions, then compute the value results with external tools (e.g., algorithms, interpreters) \cite{huang-etal-2018-using,wang2023metareasoning}.
For example, the constant expression is a commonly used IMR \cite{roy-roth-2015-solving,koncel-kedziorski-etal-2016-mawps}.
The program is another common IMR used by the current state-of-the-art (SOTA) methods \cite{chen2022program,gao2022pal,xie2023decomposition}.
Examples of these two types of IMRs are shown in Figure~\ref{fig:intermediate}.
Current methods mainly use large language models (LLMs) to generate IMRs because LLMs can use few-shot inference to generate various IMRs without training \cite{jin2023tabcot,xie2023decomposition}.


In addition to the IMRs above, previous work has also used systems of equations as IMRs \cite{roy-etal-2016-equation,heyueya2023solving}.
From an intuitive view, using equations as IMRs should be better than using programs because equations do not need to define variables before using them, leading to closer semantics to natural language questions.
An intuitive example is shown in Figure~\ref{fig:intermediate}.
However, the current method using equations as IMRs does not have better performance than that of using programs \cite{heyueya2023solving}.
So in this paper, we try to use equations as IMRs to solve the numerical reasoning task by addressing two problems:
\textbf{\textit{(1) Theoretically, how to prove that the equation is an IMR with higher generation accuracy than programs}};
\textbf{\textit{(2) Empirically, how to improve the generation accuracy of equations with LLMs}}.

For the first problem, we present and prove a proposition to compare the generation accuracy of different IMRs:
\textbf{\textit{given two IMRs, IMR\bm{$_{A}$} and IMR\bm{$_{B}$}, if IMR\bm{$_{A}$} is the subset of IMR\bm{$_{B}$}, then the accuracy in generating IMR\bm{$_{B}$} of questions theoretically surpasses IMR\bm{$_{A}$}.}}
Based on this proposition, we can prove that the accuracy of generating equations is higher than that of programs.
Because programs can be seen as equations with the restriction ``\textit{variables must be defined before being used}'', programs are a subset of equations.
Consequently, employing equations as IMRs confers theoretically higher generation accuracy than that of programs.

For the second problem, current LLMs have poor performance in generating equations \cite{heyueya2023solving}.
We assume this is because current LLMs are primarily pre-trained on constant expressions and programs for numerical reasoning \cite{brown-etal-2020-Learner,chen2021evaluating}, which makes LLMs prefer to generate these two types of IMRs rather than other IMRs during few-shot inference.
This limits the numerical reasoning ability of LLMs since these two types of IMRs are not the best IMRs for this task in theory based on our proposition.
To lower the tendency of LLMs to generate constant expressions and programs, we propose our method called \textbf{B}oosting Numerical \textbf{R}eason\textbf{i}ng by \textbf{D}ecomposing the \textbf{G}eneration of \textbf{E}quations (\ourmethod).
Our method erases asking parts and decomposes questions into sub-questions, which can improve the tendency of LLMs to generate equations.

To evaluate the effectiveness of \ourmethod, we adopt experiments on GSM8K~\cite{cobbe2021training}, SVAMP~\cite{patel-etal-2021-nlp}, and Algebra~\cite{heyueya2023solving}, which are mainstream datasets of the numerical reasoning task.
\ourmethod improves 1.6\% performance over the previous SOTA results on all above datasets on average and achieves new SOTA results under the single reasoning path setting.
In addition, ablation experiments show that \ourmethod can improve the proportion of equations in generated results, which shows that our method can indeed improve the tendency of LLMs to generate equations as IMRs.
Our contribution can be summarized as follows:
\begin{itemize}
    \item
        To theoretically prove that equations have higher generation accuracy than the IMRs of the current SOTA methods, we present and prove a proposition that can theoretically compare the generation accuracy of different IMRs.
    \item
        To empirically improve the performance of LLMs in generating equations other than constant expressions and programs, we present \ourmethod, which improves the tendency of LLMs to generate equations as IMRs.
    \item
        To verify the effectiveness of \ourmethod, we conduct experiments on multiple mainstream numerical reasoning datasets, where our method achieves new SOTA results on all datasets under the single reasoning path setting.
\end{itemize}

    \section{Methodology}
In this section, we introduce our work in detail.
First, we intuitively present why we use equations as IMRs in Section~\ref{subsec:equations} as an intuitive explanation for our proposition.
Then, we theoretically present and prove a proposition that can compare the generation accuracy of different IMRs in Section~\ref{subsec:proposition}.
After that, we empirically generate the equations for the numerical reasoning task by introducing the pipeline of \ourmethod in Section~\ref{subsec:pipeline}.

\subsection{Equations as Intermediate Meaning Representation}
    \label{subsec:equations}

    \paragraph{Challenges in IMR Design}
        The previous research has shown that even with the same model architecture, generating different IMRs may lead to different performances \cite{huang-etal-2018-using,li-etal-2022-exploring-secrets}.
        Therefore, the design of IMRs also affects the accuracy of generating.
        Generally, the more restrictive rules there are on IMRs, the harder it is for the model to generate such IMRs.
        Because if IMRs have more restrictions, the model needs more reasoning steps to meet these restrictions, resulting in the semantic difference with the question, increasing the difficulty of reasoning.

    \paragraph{Comparison of Different IMRs}
        A commonly used IMR is the constant expression \cite{roy-roth-2015-solving, koncel-kedziorski-etal-2016-mawps}, which answers the numerical reasoning question with one expression only containing constant values.
        This leads to a tremendous semantic difference between the questions and the constant expressions because questions may describe many relationships between different quantities.
        To address these restrictions, previous works propose using programs, which is the IMR of the current SOTA methods \cite{chen2022program,gao2022pal,xie2023decomposition}.
        Programs have two improvements compared with constant expressions.
        First, the program permits the utilization of variables for computing other variables, extending beyond the use of constant values exclusively.
        Second, programs can use multiple statements rather than just one statement to represent the answer.
        These two improvements make the generated programs have fewer restrictions, leading to a closer semantic gap between the program and the question that makes the generation easier. 
        However, programs also have restrictions, where each variable must be defined before use.
        This also leads to a semantic difference between the IMR and the question because one question may describe the relationships of variables before giving their values.
        To solve this restriction, we propose to use the equation as IMR, where the equation allows variables to be used before their definition \cite{roy-etal-2016-equation,heyueya2023solving}. 
        Therefore, the closer semantic gap between the equation and the question makes generation easier.

    \paragraph{Discovery from IMR Comparison}
        The above analysis shows that the above IMRs with different restrictions have subset containment relationships.
        For example, programs are equations with the restriction ``\textit{variables must be defined before being used}'', where the programs are the subset of the equations.
        Constant expressions are programs with the restriction ``\textit{there is only one assignment statement with only specific values}'', where the constant expressions are the subset of the programs.
        Based on the intuitive analysis of the previous paragraph, these subset containment relationships lead to the difference in generation accuracy.
        In Section~\ref{subsec:proposition}, we summarize this discovery into a proposition to guide the design of IMRs with high generation accuracy.

\subsection{Generation Accuracy of Different Intermediate Meaning Representations}
    \label{subsec:proposition}

    In the following, we prove that generating equations has higher accuracy than generating programs, for which we present a proposition that can theoretically compare the generation accuracy of different IMRs.
    The proof of all propositions in this section can be seen in Appendix~\ref{app:proof}.
    We first propose an auxiliary proposition to prove our main proposition:
    
    \begin{proposition}
        \label{pro:hop}
        Given IMR$_A$ and IMR$_B$, $A \subseteq B$. Given one natural language question $q$. Let $N(q, x)$ denote the hop number \cite{yang-etal-2018-hotpotqa} required to generate $x$ based on $q$.
        Then $\exists b \in B, \forall a \in A, N(q, b) \leq N(q, a)$, where all $a, b$ are the IMRs of $q$.
    \end{proposition}

    The hop number in Proposition~\ref{pro:hop} can be regarded as a numerical quantification of the difficulty of generating different IMRs.
    Intuitively, if IMR$\rm_{A}$ is a subset of IMR$\rm_{B}$, it means that IMR$\rm_{A}$ has more restrictions than IMR$\rm_{B}$, so more reasoning hops are needed to convert a natural language question to the corresponding IMR$\rm_{A}$.
    With Proposition~\ref{pro:hop}, we can present the proposition to compare the generation accuracy of different IMRs:

    \begin{proposition}
        \label{pro:accuracy}
        Given IMR$_A$ and IMR$_B$, $A \subseteq B$. Given one natural language question $q$.
        Then generating IMR$_B$ of $q$ has higher accuracy than IMR$_A$.
    \end{proposition}

    Based on Proposition~\ref{pro:accuracy}, we can compare the generation accuracy of different IMRs by judging the inclusion relationship of IMRs.
    Considering the discussion in Section~\ref{subsec:equations}, programs are a subset of equations, so generating equations as IMRs has higher accuracy than generating programs in theory.
    However, Proposition~\ref{pro:accuracy} can only compare the generation accuracy of different IMRs without considering the actual application scenario.
    Apart from the type of IMR, the generation accuracy also depends on many other factors, like the model architecture.
    Empirically, we design \ourmethod in Section~\ref{subsec:pipeline} to enhance the equation generation ability of LLMs.

\begin{figure*}[t]
    \centering
    \includegraphics[width=\linewidth]{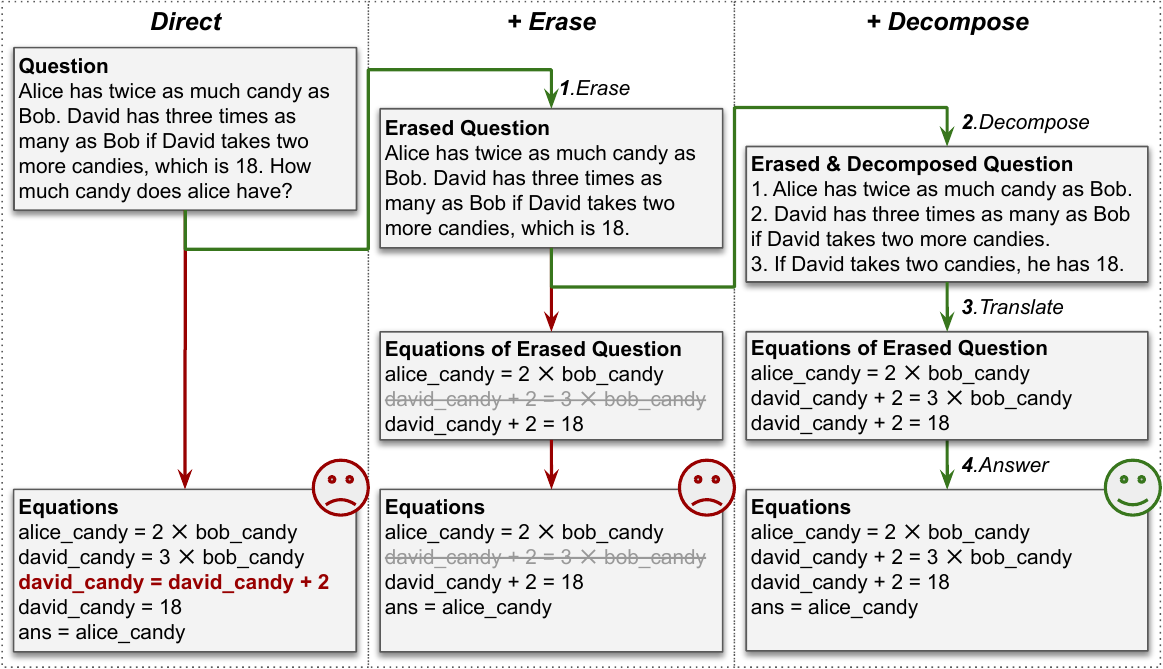}
    \caption{
        The illustration of \ourmethod under different settings of reasoning stages.
        The incorrect reasoning paths and results are annotated with \textcolor{red}{red}.
        The Direct method modifies the value of the constant unknown   ``$david\_candy$'' like the program.
        The method that only uses Erase misses the equation ``$david\_candy + 2 = 3 \times bob\_candy$'' since it pretends to translate one entire sentence into one single equation.
        The correct one is annotated with \textcolor{dark_green}{green}, which decomposes the numerical reasoning of LLMs into four stages.
        \textit{(1) Erase} the asking part of the question;
        \textit{(2) Decompose} the question into multiple sub-questions;
        \textit{(3) Translate} the sub-questions to equations;
        \textit{(4) Answer} the question based on the translated equations.
    }
    \label{fig:pipeline}
\end{figure*}

\subsection{Pipeline of \ourmethod}
    \label{subsec:pipeline}

    In the following, we introduce the pipeline of \ourmethod, which decomposes the numerical reasoning into four stages.
    The illustration of \ourmethod is shown in Figure~\ref{fig:pipeline}.
    The concrete prompt examples of \ourmethod are shown in Appendix~\ref{app:prompts}.

    \subsubsection{Erase}
        The previous research has shown that the current NLP models mainly learn the mapping between input and output formats rather than specific NLP capabilities \cite{mccoy-etal-2019-right,jawahar-etal-2019-bert,bubeck2023sparks}.
        So during the few-shot inference, current LLMs are more inclined to generate constant expressions and programs since we assume these IMRs are mainly contained in the pre-training data \cite{brown-etal-2020-Learner,chen2021evaluating}.
        However, since the pre-training data is not available as an open resource, we have been unable to validate this assumption thus far.
        To enhance the tendency of current LLMs to generate equations, we should disrupt the input format of numerical reasoning questions that LLMs have seen in the pre-training data.

        We observe that even if the asking part is erased, the remaining part can still be expressed as solvable equations.
        For example, about the question ``\textit{Alice has twice as much candy as Bob. How much candy does Alice have if Bob has 12 candies}'', after erasing the asking part ``\textit{How much candy does Alice have}'', the rest part can still be listed as equations ``$alice\_candy = 2 \times bob\_candy$, $bob\_candy = 12$''.
        To guide the LLMs in generating equations, we erase the asking part of each question.
        This disrupts the input format, as observed in the pre-training data, and reduces the tendency to generate constant expressions or programs.

    \subsubsection{Decompose}
        When generating equations based on questions, LLMs may translate one sentence into one single equation, resulting in missing intermediate information and unsolvable equations.
        Take the results only with Erase stage as an example in Figure~\ref{fig:pipeline}, LLMs directly translate ``\textit{David has three times as many as Bob if David takes two more candies, which is 18}'' into ``$david\_candy + 2 = 18$'' while ignoring the information ``\textit{David has three times as many as Bob if David takes two more candies}''.
        To address this issue, we try to decompose the question into sub-questions before generating the equations.
        With the decomposed sub-questions, LLMs can generate equations based on finer-grained information, thus alleviating missing information.
    
    \subsubsection{Translate}
        \label{subsubsec:translate}
        In this stage, we use both the erased question and the decomposed sub-questions as the input to generate the corresponding equations with LLMs, which complement each other to generate the complete equations.
        However, the Translation stage may generate unsolvable equations where no set of values that satisfies all the equations simultaneously.
        This makes it to be unable to get the value of each quantity to calculate the answer in the next stage.
        When unsolvable equations are generated, we reset the result and allow the subsequent stage to regenerate all equations.

    \subsubsection{Answer} 
        \begin{algorithm}
            \begin{algorithmic}[1]
                \item[] {\bfseries Input:}
                    Natural language question $question$,
                    translated equations $equations$,
                    number of retries limit $retry$.
                \item[] {\bfseries Output:}
                    Equations of the answer.
                \STATE $temperature = 0$
                \FOR{$i = 1$ to $retry$}
                    \STATE $ans$ = answer($question$, $equations$, $temperature$)
                    \IF{is\_solvable($equations$, $ans$)}
                        \RETURN $ans$
                    \ELSE
                        \STATE $temperature$ += $0.1$
                    \ENDIF
                \ENDFOR
                \RETURN None
            \end{algorithmic}
            \caption{Generate Answer Equation}
            \label{alg:algorithm}
        \end{algorithm}

        We need to collectively compute the answer to the numerical reasoning question, but the Translation stage only computes the value of each quantity separately.
        Concretely, in this stage, we use the original questions and the translated equations as input and output the final answer equation.
        The process of this stage is shown in Algorithm~\ref{alg:algorithm}.
        LLMs can use the unknown in the equations of the Translate equations to represent the answer equations.
        That is because the constant, variable, and operation in the equation all have certain semantic information, and LLMs excel at understanding and generating meaningful semantic terms.
        If the equations are unsolvable, we can not get the answer.
        So we let the LLMs regenerate the results if the generated equations are unsolvable and gradually increase the generation temperature until the results are solvable.

    \section{Experiments}
In this section, we adopt experiments to verify the effectiveness of our method.
First, we give the setup information of our experiments in Section~\ref{subsec:experiment_setup}.
Secondly, we present the main experiment results to demonstrate that our method can improve the numerical reasoning ability of LLMs in Section~\ref{subsec:main_experiment}.
Then, to prove the effectiveness of each stage of \ourmethod, we give the ablation experiments of each stage in Section~\ref{subsec:ablation_experiment}.
After that, we answer several research equations about the experiment results of \ourmethod to shed light on future research in Section~\ref{subsec:experiment_analysis}.
Lastly, we present a case study to better understand how \ourmethod improves the numerical reasoning ability of LLMs.

\subsection{Experiment Setup}
    \label{subsec:experiment_setup}

    \paragraph{Dataset}
        We adopt \ourmethod to GSM8K \cite{cobbe2021training}, SVAMP \cite{patel-etal-2021-nlp}, and Algebra \cite{heyueya2023solving}, which are widely used numerical reasoning datasets.
        GSM8K is the dataset consisting of grade school math questions, which require 2-8 steps of reasoning and use basic arithmetic operations ($+-\times\div$) to get answers.
        SVAMP contains $1,000$ math questions manually selected from the existing numerical reasoning datasets, which are also solved with basic arithmetic operations.
        Different from the above datasets, Algebra is a dataset containing more algebra questions, leading to a higher proportion of cases that need to be solved with equations, which can reflect the ability of the model to translate the questions into the corresponding algebraic equations.

    \paragraph{Metric}
        Following the previous work~\cite{chen2022program,gao2022pal}, we use the exact match (EM) as the evaluation metric of our work and consider the prediction is equal to the ground truth if their relative difference is below $10^{-3}$ because of the round-off error.

    \paragraph{Model}
        We use Codex~\cite{chen2021evaluating} and GPT3.5~\footnote{\url{https://platform.openai.com/docs/models/gpt-3-5}} as our experimental LLMs, which belong to the most widely used LLMs.
        Codex is an advanced model capable of translating natural language instructions into code across various programming languages.
        GPT3.5 is the model improved on GPT-3 \cite{brown-etal-2020-Learner} and can handle both natural language and code.

    \paragraph{Implement Detail}
        For the equation generation, we use the Azure OpenAI API of \texttt{code-davinci-002} and \texttt{gpt-3.5-turbo} for our experiments \footnote{\url{https://azure.microsoft.com/en-us/products/cognitive-services/openai-service}}.
        The settings of API during inference can be seen in Appendix~\ref{app:parameters}.
        We use 5-8 shots for all the experimental datasets with different hardness.
        Notably, we do not carry the experiments in multiple reasoning paths setting (e.g., self-consistency~\cite{wang2023selfconsistency}, decomposition~\cite{xie2023decomposition}) since the cost is significantly higher, amounting to 40 times that of the single path.
        Furthermore, the results we have presented in Table~\ref{tab:single_main_result} vividly illustrate the effectiveness of equations compared to other IMRs.
        We present a way of applying the multiple reasoning paths setting to \ourmethod in Appendix~\ref{app:multi_reasoning}.
        For equation solving, we employ sympy \cite{meurer-etal-2017-sympy} to solve generated equations, which is a Python package for symbolic mathematics.

\subsection{Main Result}
    \label{subsec:main_experiment}

    \begin{table*}[htp]
        \centering
        \caption{
            The exact match under different datasets and models with different prompt methods.
            The results of our method are averaged over five runs.
            The best results of different datasets are annotated in \textbf{bold}.
        }
        \begin{tabular}{llccc}
            \toprule
            \textbf{Model} & \textbf{Method} & \textbf{GSM8K} & \textbf{SVAMP} & \textbf{Algebra} \\
            \midrule
            \multirow{6}{*}{\texttt{code-davinci-002}}
             & CoT~\cite{wei2022chain} & $65.6$ & $74.8$ & $47.9$ \\
             & Tab-CoT~\cite{jin2023tabcot} & $61.6$ & $82.9$ & $-$ \\
             & Declarative~\cite{heyueya2023solving} & $69.4$ & $-$ & $76.3$ \\
             & PoT~\cite{chen2022program} & $71.6$ & $85.2$ & $-$ \\
             & PAL~\cite{gao2022pal} & $72.0$ & $79.4$ & $56.2$ \\
            \cmidrule{2-5}
             & \ourmethod & \bm{$74.2 \pm 0.4$} & \bm{$86.1 \pm 0.5$} & \bm{$78.5 \pm 1.7$} \\
            \midrule
            \multirow{3}{*}{\texttt{gpt-3.5-turbo}}
             & CoT (Our runs) & 76.5 & 80.8 & 53.6 \\
             & PoT (Our runs) & 74.8 & 79.3 & 64.0 \\
            \cmidrule{2-5}
             & \ourmethod & \bm{$77.2 \pm 0.4$} & \bm{$82.3 \pm 0.6$} & \bm{$82.0 \pm 0.9$} \\
            \bottomrule
        \end{tabular}
        \label{tab:single_main_result}
    \end{table*}

    We list the main results in Table~\ref{tab:single_main_result} and introduce the compared systems of our experiments in Appendix~\ref{app:systems}.
    We can see that \ourmethod brings robust performance improvement, leading to new SOTA results on all datasets and models under the single reasoning path setting, which proves the effectiveness of our method.

    \paragraph{GSM8K}
        Compared with the results of the previous SOTA methods, \ourmethod consistently surpasses the previous methods using programs as IMRs (PoT, PAL), empirically proving that equations are a better IMR than programs and the correctness of Proposition~\ref{pro:accuracy} to a certain extent.

    \paragraph{SVAMP}
        \ourmethod also achieves the new SOTA result with $0.9\%$ improvement.
        Notably, we can observe that the performance boost brought by \ourmethod in SVAMP is not as valid as GSM8K.
        That is because our method can only address the partial hard case, where future work should enhance the understanding ability of LLMs to relationships of quantities further than simply changing IMRs.

    \paragraph{Algebra}
        \ourmethod brings $1.7\%$ improvement compared with the previous SOTA result.
        However, our method is less robust compared with other datasets, with a fluctuation of more than $1.7\%$.
        It is because Algebra is smaller than the other two datasets with only $222$ questions, resulting in more obvious performance fluctuations.

\subsection{Ablation Study}
    \label{subsec:ablation_experiment}

    \subsubsection{Answer Generation}
        \begin{table}[t]
            \centering
            \caption{
                The ablation experiment results in Erase and Decompose stages on GSM8K and SVAMP of different models.
                The first line of each model denotes using \ourmethod.
            }
            \begin{tabular}{lllll}
                \toprule
                \textbf{Model} & \textbf{Method} & \textbf{GSM8K} & \textbf{SVAMP} & \textbf{Algebra} \\
                \midrule
                \multirow{3}{*}{\texttt{code-davinci-002}}
                 & \ourmethod & $74.2$ & $86.1$ & $78.5$ \\
                 & - Erase & $69.0(-5.2)$ & $86.0(-0.1)$ & $72.1(-6.4)$ \\
                 & - Decompose & $68.2(-6.0)$ & $85.6(-0.5)$ & $77.8(-0.7)$ \\
                \midrule
                \multirow{3}{*}{\texttt{gpt-3.5-turbo}}
                 & \ourmethod & $77.0$ & $82.5$ & $82.9$ \\
                 & - Erase & $76.1(-0.9)$ & $81.9(-0.6)$ & $81.5(-1.4)$ \\
                 & - Decompose & $74.2(-2.8)$ & $81.9(-0.6)$ & $79.7(-3.2)$ \\
                \bottomrule
            \end{tabular}
            \label{tab:ablation_result}
        \end{table}

        We adopt the ablation study to verify the stage effectiveness of \ourmethod.
        The experiment result is shown in Table~\ref{tab:ablation_result}, from which we can see that:
        \textit{(1)} The Erase and Decompose stages can bring improvement to all datasets, proving the effectiveness of these two stages;
        \textit{(2)} Compared with Erase, using Decompose can bring a more significant improvement on GSM8K and SVAMP since Decompose is also effective for questions that do not require equation solving, while Erase is mainly designed for questions using equations;
        \textit{(3)} Compared with GSM8K, the improvement in SVAMP is less obvious because the questions of SVAMP are much simpler than GSM8K, and most questions not solved by previous methods require enhancing abilities other than numerical reasoning, so the improvement brought by our method is not significant.

    \subsubsection{Equation Generation}
        \begin{table}[t]
            \centering
            \caption{
                The equations generated with and without Erase stage.
                \textbf{Correct} denotes the correct cases, and \textbf{Total} denotes the total cases.
                $\Delta$ denotes the ratio of the equations increased after using Erase to that without Erase.
            }
            \begin{tabular}{llcccccc}
                \toprule
                \multirow{2}{*}{\textbf{Model}} & \multirow{2}{*}{\textbf{Method}} & \multicolumn{2}{c}{\textbf{GSM8K}} & \multicolumn{2}{c}{\textbf{SVAMP}} & \multicolumn{2}{c}{\textbf{Algebra}} \\
                \cmidrule(r){3-4} \cmidrule(r){5-6} \cmidrule(r){7-8}
                 & & \textbf{Correct} & \textbf{Total} & \textbf{Correct} & \textbf{Total} & \textbf{Correct} & \textbf{Total} \\
                \midrule
                \multirow{3}{*}{\texttt{code-davinci-002}}
                 & \ourmethod & $83$ & $108$ & $49$ & $57$ & $68$ & $80$ \\
                 & - Erase & $61$ & $89$ & $44$ & $52$ & $65$ & $77$ \\
                \cmidrule{2-8}
                 & $\Delta$ & $36.1\%$ & $21.3\%$ & $11.4\%$ & $9.6\%$ & $4.6\%$ & $3.9\%$ \\
                \midrule
                \multirow{3}{*}{\texttt{gpt-3.5-turbo}}
                 & \ourmethod & $93$ & $139$ & $56$ & $65$ & $77$ & $86$ \\
                 & - Erase & $65$ & $96$ & $43$ & $51$ & $75$ & $81$ \\
                \cmidrule{2-8}
                 & $\Delta$ & $43.1\%$ & $44.8\%$ & $30.2\%$ & $27.5\%$ & $2.7\%$ & $6.2\%$ \\
                \bottomrule
            \end{tabular}
            \label{tab:ablation_equation}
        \end{table}
    
        To verify whether the Erase stage can improve the tendency of LLMs to generate equations, we count the number of generated equations in the results with and without Erase.
        The experiment results are shown in Table~\ref{tab:ablation_equation}.
        From the results, we can see that:
        \textit{(1)} On each dataset, the numbers of results using equations on both correct and total cases have increased, proving that the Erase stage indeed improves the tendency of LLMs to generate equations;
        \textit{(2)} Compared with GSM8K, the improvement of the Erase stage in Algebra is not obvious because Algebra mainly includes algebra questions, which can guide to generate algebra equations by questions themselves.

\subsection{Analysis}
    \label{subsec:experiment_analysis}

    In the following, we conduct an analysis of the experiment results to answer several research questions.

    \begin{githubquote}
        \textbf{RQ1}. Does the prompt of \ourmethod can significantly affect the performance? \textbf{Yes} \
    \end{githubquote}
        \begin{figure*}[t]
            \centering
            \begin{tikzpicture}
                \begin{axis} [
                    ybar,
                    height=1.2in, 
                    bar width=0.4cm,
                    width=0.95\linewidth,
                    scale only axis,
                    ymin = 60, 
                    ymax = 90,
                    xticklabel style = {font=\small},
                    yticklabels=\empty,
                    axis x line*=bottom,
                    hide y axis,
                    enlarge x limits=0.2,
                    xtick={2-shots, 4-shots, 6-shots, 8-shots},
                    symbolic x coords={2-shots, 4-shots, 6-shots, 8-shots},
                    xticklabel style = {font=\small,yshift=0.5ex},
                    nodes near coords,
                    nodes near coords align={vertical},
                    every node near coord/.append style={font=\tiny},
                    legend style={
                        at={(0,1.0)},
                        anchor=north west,
                        legend columns=-1
                    },
                ]
                    \addplot coordinates {
                    (2-shots, 72.3) (4-shots, 72.0) (6-shots, 73.7) (8-shots, 75.0)
                    };
                    \addplot coordinates {
                    (2-shots, 70.1) (4-shots, 72.9) (6-shots, 74.6) (8-shots, 75.8)
                    };
                    \addplot coordinates {
                    (2-shots, 71.6) (4-shots, 71.6) (6-shots, 74.3) (8-shots, 77.6)
                    };
                \end{axis}
            \end{tikzpicture}
            \caption{
                The exact match of inference with different shots on GSM8K run three times using \texttt{gpt-3.5-turbo}.
                Different color denotes the result of the different run.
            }
            \label{fig:prompt_analysis}
        \end{figure*}
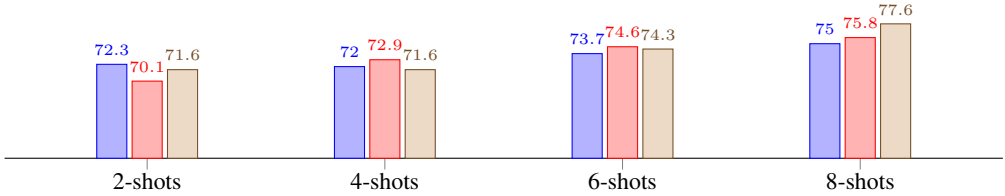

        To study the impact of different prompts on \ourmethod, we write a total of $15$ samples and randomly select \{$2$, $4$, $6$, $8$\} shots from them as prompts to run three times.
        The experiment results are shown in Figure~\ref{fig:prompt_analysis}.
        From the results, we can see that as the number of shots in the prompt increases, the performance of our method gradually improves.
        Besides, as the number of shots increases, the variance of the results is gradually decreasing, indicating that the robustness is also gradually improving.

    \begin{githubquote}
        \textbf{RQ2}. Does \ourmethod increase the tendency of LLMs to generate equations? \textbf{Yes}
    \end{githubquote}
        \begin{table}[t]
            \centering
            \caption{
                Among the cases where PoT is incorrect while \ourmethod is correct, the proportion of cases using equations.
            }
            \begin{tabular}{lccc}
                \toprule
                \textbf{Model} & \textbf{GSM8K} & \textbf{SVAMP} & \textbf{Algebra} \\
                \midrule
                \texttt{code-davinci-002} & $20.8\%$ & $15.3\%$ & $47.8\%$ \\
                \texttt{gpt-3.5-turbo} & $41.7\%$ & $10.0\%$ & $59.6\%$ \\
                \bottomrule
            \end{tabular}
            \label{tab:proportion_results}
        \end{table}

        To evaluate whether \ourmethod enhances the numerical reasoning ability of LLMs by generating equations, we select the cases where PoT is incorrect while our method is correct and see if each case generates equations as IMRs.
        The proportions of generating equations of different methods and datasets are shown in Table~\ref{tab:proportion_results}.
        Based on the results, we can see that:
        \textit{(1)} On all models and all datasets, the performance improvement has equations involved, and it is most significant in Algebra, where about half of the improvement uses equations;
        \textit{(2)} The performance improvement is not all brought by using equations as IMRs because the results of LLMs are not robust, and many questions in the part that does not use equations may be solved in the results of other runs of PoT;
        \textit{(3)} The improvement brought by using equations on SVAMP is not significant because the questions of this dataset are relatively simple, and most of them can be solved without equations.

    \begin{githubquote}
        \textbf{RQ3}. Does \ourmethod enhance the performance on all difficulty levels? \textbf{Yes}
    \end{githubquote}
        \begin{table}[t]
            \centering
            \caption{
                The exact match of questions with different equations steps of \texttt{code-davinci-002}.
                \#Steps denotes the number of answer equations.
            }
            \begin{tabular}{lcccccc}
                \toprule
                \multirow{2}{*}{\textbf{\#Steps}} & \multicolumn{3}{c}{\textbf{GSM8K}} & \multicolumn{3}{c}{\textbf{Algebra}} \\
                \cmidrule(r){2-4} \cmidrule(r){5-7}
                 & \textbf{PoT} & \textbf{\ourmethod} & \textbf{$\Delta$} & \textbf{PoT} & \textbf{\ourmethod} & \textbf{$\Delta$} \\
                \midrule
                $\leq4$ & $71.1$ & $78.6$ & $7.5$ & $49.5$ & $80.5$ & $31.0$ \\
                $[5, 6]$ & $71.6$ & $78.3$ & $6.7$ & $20.0$ & $60.0$ & $40.0$ \\
                $\geq7$ & $68.5$ & $70.0$ & $1.5$ & $5.9$ & $58.8$ & $52.9$ \\
                \bottomrule
            \end{tabular}
            \label{tab:complex_results}
        \end{table}

        We analyze the effectiveness of our method for solving questions from various difficulty levels, each categorized by the number of statements or answer equations.
        The performance improvement of our method compared with PoT is shown in Table~\ref{tab:complex_results}, which shows that:
        \textit{(1)} Our method can bring performance improvement of questions under different complexes, especially on the Algebra dataset, improvement is more than 30\%, which proves the effectiveness of our method under different complexes;
        \textit{(2)} On the GSM8K dataset, the improvement of more complex questions is not obvious, showing that \ourmethod brings relatively small performance improvement when dealing with more complex questions.

    \begin{githubquote}
        \textbf{RQ4}. At which stages does \ourmethod mainly make mistakes? \textbf{Translate and Answer Stages}
    \end{githubquote}
        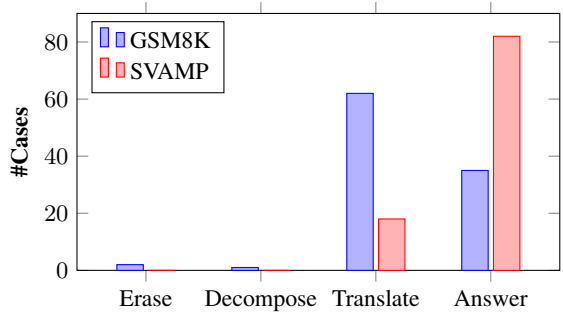
\begin{figure}[t]
            \centering
            \begin{tikzpicture}
                \small
                \begin{axis}[
                    ybar,
                    ymin=0,
                    ymax=90,
                    height=5cm,
                    ylabel={\textbf{\#Cases}},
                    legend style={cells={anchor=west}, legend pos=north west},
                    symbolic x coords={Erase, Decompose, Translate, Answer},
                    xtick=data,
                    width=8cm,
                    enlarge x limits=0.2,
                ]
                    \addplot+[ybar] plot coordinates {(Erase,2) (Decompose,1) (Translate,62) (Answer,35)};
                    \addplot+[ybar] plot coordinates {(Erase,0) (Decompose,0) (Translate,18) (Answer,82)};
                    \legend{GSM8K, SVAMP}
                \end{axis}
            \end{tikzpicture}
            \caption{
                The number of bad cases with \texttt{code-davinci-002} in different stages of GSM8K and SVAMP.
            }
            \label{fig:error_analysis}
        \end{figure}

        In order to investigate the main sources of error in \ourmethod, we count the number of bad cases where errors occurred in different stages.
        We randomly select one hundred bad cases of GSM8K and SVAMP, then manually classify them according to the stages where the error occurred.
        The results are shown in Figure~\ref{fig:error_analysis}.

        Based on the results, we can draw the conclusion that:
        \textit{(1)} LLMs make very few mistakes in the Erase and Decompose stages. These stages rely on understanding the meaning of questions and do not require reasoning, leading to better performance.
        \textit{(2)} Most mistakes occur during the Translate and Answer stages. This indicates that current LLMs still hard to convert regular language into equations. Even after breaking down the question, they still get mixed up between the meanings of questions and equations.
        \textit{(3)} GSM8K and SVAMP show different kinds of errors. GSM8K struggles because its questions involve complex quantity relationships that are tough to turn into equations. SVAMP, on the other hand, faces difficulty in figuring out which quantity the question is asking about.

\subsection{Case Study}
    \label{subsec:case_study}

    \begin{figure}[t]
        \centering
        \includegraphics[width=0.7\linewidth]{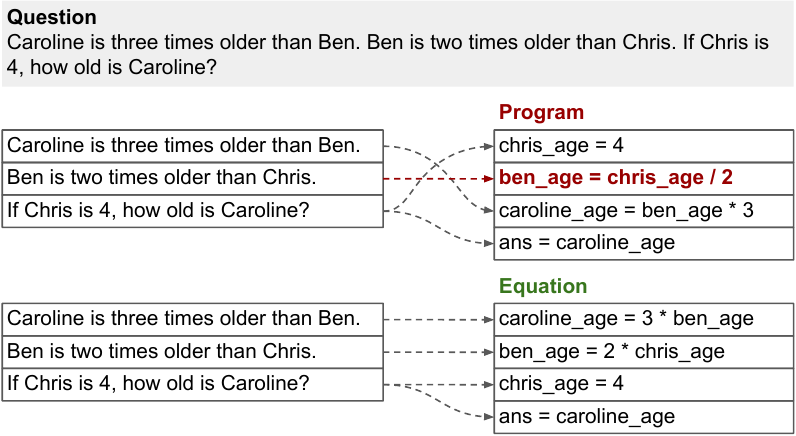}
        \caption{
            The results of PoT and \ourmethod of a case in GSM8K.
            The above part is the result of PoT, and the below part is \ourmethod.
            The error steps and results are annotated in \textcolor{red}{red}, and the correct results are annotated in \textcolor{dark_green}{green}.
        }
        \label{fig:case_study}
    \end{figure}

    To better understand how \ourmethod improves the numerical reasoning performance of LLMs, we present a case study in Figure~\ref{fig:case_study}.
    Since the program needs to set up variables before using them, when tackling the question in Figure~\ref{fig:case_study}, the first step is understanding how different quantities are related. 
    Then, the order of each sub-question in the result needs adjustment, but this tends to have low accuracy in generation.
    In contrast, \ourmethod adopts the equation as IMR., where the generated answers precisely match each sentence of the original question, ensuring a straightforward and accurate generation process.

    \section{Related Work}
        \subsection{Numerical Reasoning with LLMs}
    The ability to perform numerical reasoning is essential for NLP models as it enables them to understand and manipulate numerical information embedded in natural language \cite{lu-etal-2023-survey}.
    Utilizing LLMs for numerical reasoning tasks has become the mainstream in current research due to their brilliant few-shot inference ability without training \cite{wei2022chain,chen2022program,xie2023decomposition}.

    Early researches guide LLMs to generate answers and the reasoning process in one step \cite{jie-lu-2023-leveraging,liu2023goat,imani-etal-2023-mathprompter}.
    For example, CoT \cite{wei2022chain} asks LLMs to generate the reasoning process with natural language to get the answer.
    Since the program can express the reasoning process more accurately than the text, PoT \cite{chen2022program} and PAL \cite{gao2022pal} ask the model to generate programs to solve numerical questions.
    Following the same reason, Tab-CoT \cite{jin2023tabcot} asks the model to generate tables to solve this task, where different columns denote the different reasoning information.

    Besides, many works also try to decompose the process of numerical reasoning into multiple steps to reduce the difficulty of model inference \cite{gaur-saunshi-2023-reasoning,li-etal-2023-making,wang2023metareasoning}.
    For example, PHP~\cite{zheng2023PHP} generates answers through multi-turn inferences, and each turn can use the answers of the previous turns as a hint.
    Decomposition~\cite{xie2023decomposition} generates multiple candidate results based on the previous steps of each reasoning step (e.g., one program statement) and keeps the best k of them like beam-search.

    However, the existing work has two limitations.
    One is that the IMRs used are not the best, and the other is that the better IMR cannot be effectively generated.
    To overcome the first problem, we present Proportion~\ref{pro:accuracy} to compare the generation accuracy of different IMRs in theory.
    To address the second problem, we present \ourmethod, which can enhance the equation generation of LLMs.

\subsection{Reasoning with Intermediate Meaning Representation}
    In the field of NLP, a common way to reduce the difficulty of reasoning is to generate answers with an intermediate meaning representation (IMR) \cite{gan-etal-2021-natural-sql,nie-etal-2022-graphq,paul2023refiner}.
    Such methods first generate IMRs of questions and then use external tools (e.g., algorithms, interpreters) to generate the answer results based on IMRs.
    Since generating the answer from IMRs is deterministic, the main bottleneck is how to generate IMRs of questions.

    The most widely studied IMRs are the semantic representation language \cite{kamath2019survey}.
    Previous works have designed many semantic representation languages to represent natural language sentences, such as AST \cite{ASTImpl2003ast} and AMR \cite{banarescu-etal-2013-abstract}.
    By converting into these languages, then applying the converted results on downstream tasks, the semantic understanding ability of the model can be effectively improved \cite{che-etal-2021-ltp}.

    Many methods also employ IMRs to solve the numerical reasoning task \cite{wang2023metareasoning,paul2023refiner}.
    Current methods have present various IMRs, such as constant expressions~\cite{roy-roth-2015-solving,koncel-kedziorski-etal-2016-mawps}, dolphin languages~\cite{huang-etal-2018-using}, domain-specific languages~\cite{chen-etal-2021-finqa} and equations~\cite{roy-etal-2016-equation,heyueya2023solving}.
    The current SOTA methods use programs as IMRs because programs have closer semantics to questions than constant expressions, and current LLMs have strong program generation capabilities \cite{chen2021evaluating,chen2022program,gao2022pal}.
    Past work did not use programs as IMRs because programs generated without LLMs performed poorly~\cite{chen2021evaluating}.

    However, most methods design IMRs based on expert experience or experimental results, affected by factors other than IMRs themselves, such as model structure and training data, leading to the high cost of IMR design.
    To theoretically compare the generation of different IMRs and lower the cost of IMR design, in this paper, we propose and prove Proposition~\ref{pro:accuracy} to guide the design of IMRs.

    \section{Conclusion}
        To theoretically and empirically advance the numerical reasoning research in the LLMs era, in this paper, we employ equations as IMRs to solve the numerical reasoning task by addressing two problems: 
        \textit{(1)} Theoretically, how to prove that the equation is an IMR with higher generation accuracy than programs;
        \textit{(2)} Empirically, how to improve the generation accuracy of equations with LLMs.
        For the first problem, we present and prove a proposition to compare the generation accuracy of different IMRs in theory.
        For the second problem, we present \ourmethod to enhance the equation generation of LLMs by reducing the tendency of generating constant expressions and programs and decomposing questions.
        To evaluate \ourmethod, we conduct experiments across three datasets: GSM8K, SVAMP, and Algebra. 
        Compared to the previous SOTA results, our method has increased average performance by 1.6\%, setting new SOTA performance across all the datasets under the single reasoning path setting. 
        Moreover, ablation experiments show that using \ourmethod can enhance the tendency of LLMs to generate equations as IMRs, proving that our method can improve the ability of LLMs to generate IMRs outside of constant expressions and programs.

    \newpage
    \bibliographystyle{plainnat}
    \bibliography{neurips_2020}

    \newpage
    \appendix
        \section{Proof of Proposition}
    \label{app:proof}

    We present the proof of the propositions in Section~\ref{subsec:proposition}. \\
    First, we prove Proposition~\ref{pro:hop}:

    \begin{proof}
        We first arbitrarily choose \( b \in B \). For any \( a \in A \), we have:
        \begin{enumerate}
            \item If \( N(q, b) \leq N(q, a) \), the proposition holds.
            \item If \( N(q, b) > N(q, a) \), because \( a \in A \subseteq B \), \( a \) is also the IMR B of \( q \). Thus, we can set \( b = a \), resulting in \( N(q, b) = N(q, a) \). The proposition still holds.
        \end{enumerate}
        In conclusion, $\exists b \in B, \forall a \in A, N(q, b) \leq N(q, a)$.
    \end{proof}

    With Proposition~\ref{pro:hop}, we can prove Proposition~\ref{pro:accuracy}:

    \begin{proof}
        Because $A \subseteq B$, based on Proposition~\ref{pro:hop}, $\exists b \in B, \forall a \in A, N(q, b) \leq N(q, a)$.
        Since the bigger of $N(q, x)$, the lower accuracy to generate $x$ based on $q$ \cite{yang-etal-2018-hotpotqa,jansen-2018-multi,min-etal-2019-multi}, generating IMR $B$ of $q$ has higher accuracy than IMR $A$.
    \end{proof}

\section{Examples of Prompts}
    \label{app:prompts}

    \begin{table}[t]
        \centering
        \caption{Prompts of the Erase stage of \ourmethod}
        \begin{tabular}{p{0.9\linewidth}}
            \toprule
            \textbf{\textit{\underline{Erase}}} \\
            Erase the asking part of the question. \\
            \\
            Question: \\
            If Marcy works for the same company for 40 years, she gets an annual pension of \$50,000/year. Starting after 20 years, she becomes entitled to 5\% of the value of the pension per year. If she quits after 30 years, what will her annual pension be? \\
            Erased: \\
            If Marcy works for the same company for 40 years, she gets an annual pension of \$50,000/year. Starting after 20 years, she becomes entitled to 5\% of the value of the pension per year. She quits after 30 years. \\
            \bottomrule
        \end{tabular}
        \label{tab:erase_promt}
    \end{table}

    \begin{table}[t]
        \centering
        \caption{Prompts of the Decompose stage of \ourmethod}
        \begin{tabular}{p{0.9\linewidth}}
            \toprule
            \textbf{\textit{\underline{Decompose}}} \\
            Decompose the paragraph. \\
            \\
            Paragraph: \\
            If Raymond does half as much laundry as Sarah, and Sarah does 4 times as much laundry as David, Sarah does 400 pounds of laundry. \\
            Decomposed: \\
            Raymond does half as much laundry as Sarah. \\
            Sarah does 4 times as much laundry as David. \\
            Sarah does 400 pounds of laundry. \\
            \bottomrule
        \end{tabular}
        \label{tab:decompose_prompt}
    \end{table}

    \begin{table}[t]
        \centering
        \caption{Prompts of the Translate stage of \ourmethod}
        \begin{tabular}{p{0.9\linewidth}}
            \toprule
            \textbf{\textit{\underline{Translate}}} \\
            Translate the paragraph into the corresponding system of equations. \\
            The name of the unknown should correspond to the meaning of the entity in the paragraph. \\
            Unknowns are unmodifiable constants. \\
            \\
            Paragraph: \\
            If Raymond does half as much laundry as Sarah, and Sarah does 4 times as much laundry as David, Sarah does 400 pounds of laundry. \\
            Decomposed: \\
            Raymond does half as much laundry as Sarah. \\
            Sarah does 4 times as much laundry as David. \\
            Sarah does 400 pounds of laundry. \\
            Equations: \\
            raymond\_laundry = sarah\_laundry / 2 \\
            sarah\_laundry = 4 * david\_laundry \\
            sarah\_laundry = 400 \\
            \bottomrule
        \end{tabular}
        \label{tab:translate_prompt}
    \end{table}

    \begin{table}[t]
        \centering
        \caption{Prompts of the Answer stage of \ourmethod}
        \begin{tabular}{p{0.9\linewidth}}
            \toprule
            \textbf{\textit{\underline{Answer}}} \\
            Generate the answer of the question based on the unknowns in the equations. \\
            The equations must end with the unknown 'ans'. \\
            Use only addition, subtraction, multiplication, division and parentheses. \\
            \\
            Question: \\
            If Raymond does half as much laundry as Sarah, and Sarah does 4 times as much laundry as David, calculate the difference in the amount of laundry Raymond and David do if Sarah does 400 pounds of laundry. \\
            Equations: \\
            raymond\_laundry = sarah\_laundry / 2 \\
            sarah\_laundry = 4 * david\_laundry \\
            sarah\_laundry = 400 \\
            ans = raymond\_laundry - david\_laundry \\
            \bottomrule
        \end{tabular}
        \label{tab:answer_prompt}
    \end{table}

    We present the prompts of our method in Table~\ref{tab:erase_promt}, Table~\ref{tab:decompose_prompt}, Table~\ref{tab:translate_prompt} and Table~\ref{tab:answer_prompt}.
    Due to page height limitations, for each prompt, we only give its instruction and one shot, and the complete prompt can be found in the source code in \sourcecode.

\section{API Settings of Inference}
    \label{app:parameters}

    In the Answer stage, we set the retry limit as $5$.
    During the inference with API, we set top\_p as $1$, and other parameters adopt default settings.
    The API version of \texttt{code-davinci-002} is \texttt{2022-12-01}, and the version of \texttt{gpt-3.5-turbo} is \texttt{2023-03-15-preview}.

\section{Comparison Systems of Experiments}
    \label{app:systems}

    In this section, we introduce the comparison systems of our experiments in Section~\ref{subsec:main_experiment}.

    \paragraph{Chain of Thought (CoT)}
        CoT~\cite{wei2022chain} is one of the most widely used prompt methods to enhance the reasoning of LLMs.
        Instead of generating answers directly, CoT asks LLMs to generate each reasoning step using natural languages, which can help LLMs explicitly derive results to better think about the relationships between different steps.

    \paragraph{Declarative}
        Since CoT needs to convert natural language into expressions and perform calculations, the semantics of the results generated by CoT may be quite different from the original questions, increasing the reasoning difficulty.
        To solve this problem, Declarative~\cite{heyueya2023solving} converts the original natural language question into a symbolic representation.
        When generating reasoning results, LLMs are asked to generate natural language reasoning steps with the corresponding symbolic representations.
        After getting the symbolic representations, the question is answered by solving the equations corresponding to the symbolic representations, thereby reducing the difficulty of reasoning.

    \paragraph{Program Aided Language (PAL)}
        Also, in order to solve the problem that CoT requires LLMs to perform numerical calculations themselves, PAL~\cite{gao2022pal} tries to let LLMs generate programs to solve numerical reasoning questions.
        Similar to Declarative, during the generation of inference results, PAL generates natural language results with the corresponding Python program statements, then obtains the value results by executing the resulting program.

    \paragraph{Program of Thought (PoT)}
        PoT~\cite{chen2022program} is another method of using Python programs to help LLMs solve the numerical reasoning task.
        Different from PAL, PoT directly allows LLMs to generate executable programs without generating natural language explanations for each statement.
        Because this format is closer to the program data seen in the pre-training data of LLMs, they have more tendency to generate the results in this format.

    \paragraph{Tabular Chain of Thought (Tab-CoT)}
        As discussed in the Tab-CoT paper~\cite{jin2023tabcot}, the above methods need to design different prompts for different domains, lacking generalization.
        To address this problem, Tab-CoT proposes to organize the steps of reasoning by generating tables and using different columns to represent different information on reasoning steps.
        In this way, LLMs show strong zero-shot inference performance.

\section{\ourmethod with Multiple Reasoning Paths}
    \label{app:multi_reasoning}

    \begin{table}[t]
        \centering
        \caption{
            The results of \texttt{code-davinci-002} on GSM8K with different prompts of multiple reasoning paths.
            \textbf{API Time} denotes the call times of API required for each case.
        }
        \begin{tabular}{lcc}
            \toprule
            \textbf{Prompt} & \textbf{EM} & \textbf{API Time} \\
            \midrule
            CoT w. Self-Consistency~\cite{wang2023selfconsistency} & $78.0$ & $40$ \\
            PoT w. Self-Consistency~\cite{chen2022program} & $80.0$ & $40$ \\
            PAL w. Self-Consistency~\cite{gao2022pal} & $80.4$ & $40$ \\
            CoT w. Diverse~\cite{li-etal-2023-diverse} & $82.3$ & $100$ \\
            PAL w. Decomposition~\cite{xie2023decomposition} & $85.5$ & $\approx 600$ \\
            \midrule
            \ourmethod w. Self-Consistency ($128$ examples) & $84.3$ & $38$ \\
            \bottomrule
        \end{tabular}
        \label{tab:my_label}
    \end{table}

    In the following, we discuss how to apply \ourmethod with multiple reasoning paths.
    Since the error cases in Erase and Decompose Stages are almost zero, as shown in Figure~\ref{fig:error_analysis}, we mainly discuss how to use multiple reasoning paths to improve the performance of equation generation.
    We generate k systems of equations in the Translate stage, then generate answers for each system of equations in the Answer Stage.
    We vote on the results and select the value with the highest votes as the final result.
    Due to the high overhead of calling the API, we only randomly sampled 128 examples for preliminary experiments.
    For each case, we generate $16$ systems of equations in the Translate stage.
    The experimental results are shown in Table.
    Although our method only achieves a performance close to Decomposition, our method calls a smaller number of APIs, leading to a significantly reduced running overhead.

\end{document}